\documentclass[conference]{IEEEtran}
\usepackage{times}

\usepackage[numbers]{natbib}
\usepackage{multicol}
\usepackage[bookmarks=true]{hyperref}
\usepackage{graphicx}
\usepackage{mathtools}
\usepackage[longend, ruled, vlined, noend]{algorithm2e}
\usepackage{subcaption}
\usepackage{color}
\usepackage{amsmath}
\usepackage{amsfonts}
\usepackage{amssymb}
\usepackage[normalem]{ulem}

\newcommand{\threshold}{\epsilon}
\newcommand{\simplex}{\mathbb{S}}
\newcommand{\sd}{\mathbf{v}}
\newcommand{\origin}{\mathbf{O}}
\newcommand{\witness}{\mathbf{w}}

\newcommand{\bP}{\mathbf{P}}
\newcommand{\bQ}{\mathbf{Q}}
\newcommand{\bp}{\mathbf{p}}
\newcommand{\bq}{\mathbf{q}}
\newcommand{\bc}{\mathbf{c}}
\newcommand{\supportFun}{support}
\newcommand{\distanceFun}{distance}

\begin{document}

\title{High Precision Real Time Collision Detection}

\author{Author Names Omitted for Anonymous Review. Paper-ID 3}

\author{\IEEEauthorblockN{Alexandre Coulombe}
\IEEEauthorblockA{Department of Electrical and Computer Engineering,\\
 McGill University, Montreal, Canada\\
 Email: alexandre.coulombe@mail.mcgill.ca\\}
\and
\IEEEauthorblockN{Hsiu-Chin Lin}
\IEEEauthorblockA{School of Computer Science,\\
 Department of Electrical and Computer Engineering,\\
 McGill University, Montreal, Canada\\
 Email: hsiu-chin.lin@cs.mcgill.ca\\}}

\maketitle

\begin{abstract}
    Collision detection and collision avoidance are essential components in these systems for safe human-robot interactions.
    Robotics systems that can work "out-of-the-box" without excessive amount of installation and calibration from the experts is highly ideal.
    For this, we propose a generic, high precision, collision detect system that only requires the unified robot description format (URDF) and is capable of running in real time.
    We extended the Gilbert-Johnson-Keerthi (GJK) algorithm by utilizing a geometrical approach to determine the distance between each rigid body in the environment and check for collisions. The proposed system's performance is shown by checking the self-collision of the KUKA LBR iiwa 7 R800 and the Mecademic Meca500. The performance is compared to the Flexible Collision Library (FCL). 
\end{abstract}

\IEEEpeerreviewmaketitle

\section{Introduction}
\label{sec:intro}
    \noindent
    With the arrival of Industry 4.0, an increase in robotic systems has emerged in manufacturing. Collision detection and collision avoidance are essential components in these systems for safe human-robot interactions~\cite{lin2018projected}. 
    In order to guarantee the safety of the users and the robots, these components need to work in real time (i.e., the computations must be performed within one millisecond). 
    However, these systems can be very complex to install and require experts to calibrate and operate.
    Therefore, robotics systems that follow an "out-of-the-box" convention is highly preferable.
    
    Many approaches use geometric volumes to model the robot and detect collisions between objects~\cite{ericson2004real}. Models such as Bounding Spheres~\cite{wu1992linear}, Axis-Aligned Bounding Boxes (AABB)~\cite{bergen1997efficient}, Oriented Bounding Boxes (OBB)~\cite{chang2010efficient} and Bounding Cylinders~\cite{ketchel2006collision} are used as an abstraction of the robot's true model and provide fast approximations. However, they achieve this at the expense of conserving the true form of the robot, losing the ability to get the true distances between objects.
    
    Some methods perform collision detection by using convex hulls that represent the true form of the robot, such as the Gilbert-Johnson-Keerthi (GJK) distance algorithm~\cite{gilbert1988GJK} which can also find the distance between the convex hulls. Improvements have been applied to this method, such as adding recursion to reduce computation~\cite{ong1997GJK},  finding the penetration distance of objects that are intersecting~\cite{cameron1997GJK}, and making the algorithm more numerically stable~\cite{montanari2017GJKSV}.

    The Flexible Collision Library (FCL) is an open-source C++ collision detection library~\cite{pan2012FCL} that features many collision detection algorithms and is integrated in the Robot Operating System (ROS). However, FCL is too slow in calculating the distance between meshes, making it unsuitable for real-time use. Also, a lot of development time and effort needs to be invested to use FCL and ROS on a robot.

    In this paper, we propose a collision detection method that is capable of being used "out-of-the-box", run in real time, and, in addition to verifying collisions between objects, get the distance between objects in the modelled environment. We also propose a variant of the GJK distance sub-algorithm using a geometrical approach. The method is generic, and only requires the robot's unified robot description format (URDF) and the objects' meshes with their respective transformations. Background is provided in Section~\ref{sec:related-work}, our proposed method is explored in Section~\ref{sec:proposed-method}, and the method is compared to the Flexible Collision Library (FCL)~\cite{pan2012FCL} in Section~\ref{sec:results}.
    
    \begin{figure}[t]
        \centering
        \includegraphics[scale=0.3]{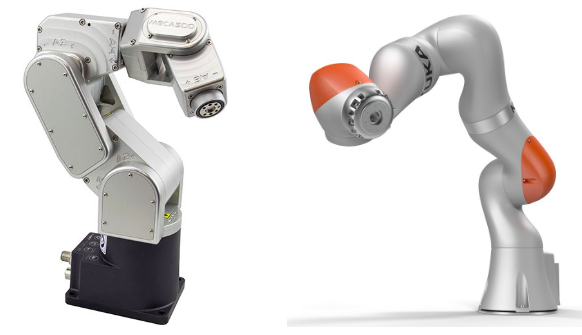}
        \caption{\centering The KUKA LBR iiwa (right) and the Mecademic Meca500 (left) were examples for collision detection.}
        \label{fig:robot-experiments}
        \vskip -0.5cm
    \end{figure}

\section{Related Work}
\label{sec:related-work}

\noindent
A typical approach for collision detection is to describe each link of the robot as a convex object and calculate the distance between each pair of objects. In this section, we provide a brief summary to distance querying between convex objects.

\subsection{Minkowski difference for distance query}
\noindent
    Given two objects defined by $\bP$ and $\bQ$, the sets of points composing them respectively, the Minkowski difference between them is defined as:
    \begin{equation}
        \bP \ominus \bQ = \{ \bp-\bq : \forall \bp \in \bP , \forall \bq \in \bQ \}
    \end{equation}
    If the origin $\origin$ is within the convex hull formed by $\bP \ominus \bQ$, then $\bP$ and $\bQ$ intersect. Otherwise, the shortest distance between $\bP$ and $\bQ$, denoted $d_{\bP,\bQ}$, is equivalent to the shortest distance between $\bP\ominus \bQ$ and $\origin$.
    \begin{equation}
    \begin{aligned}
       d_{\bP,\bQ} & = \min \{ ||\bp-\bq|| : \forall \bp \in \bP , \forall \bq \in \bQ \} \\
        & = \min \{ ||\bc -\origin||: \forall \bc \in \{\bP \ominus \bQ \} \}
    \end{aligned}
    \label{equ:mikowski-distance}
    \end{equation}
   
\subsection{Gilbert-Johnson-Keerthi Algorithm}
\label{sec:GJK}
\begin{figure}[h]
        \centering
        \includegraphics[scale=0.5]{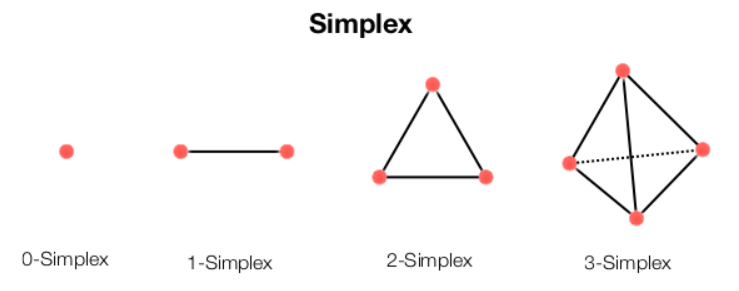}
        \caption{\centering The possible simplexes that can be used in three dimensions for the GJK algorithm}
        \label{fig:simplexes}
        \vskip -0.5cm
    \end{figure}

\noindent
    Evaluating the entire Minkowski difference in Eq.~\ref{equ:mikowski-distance} is infeasible in real time; alternatively, the GJK algorithm iteratively approximates the convex hull of $\bP \ominus \bQ$ as a {\em simplex} \cite{gilbert1988GJK}. In $\mathbb{R}^3$, the simplexes are the point, line, triangle and tetrahedron as seen in Fig.~\ref{fig:simplexes}.
    
\begin{figure}[h]
        \centering
        \begin{subfigure}{.45\linewidth}
            \centering
            \includegraphics[width=0.8\linewidth]{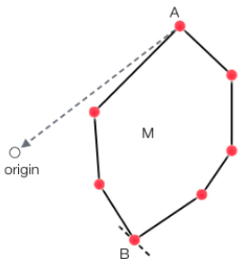}
            \caption{\centering Initially, the witness point is A and $\simplex$ is A. The direction $\sd=\overrightarrow{AO}$.}
        \end{subfigure}
        \begin{subfigure}{.45\linewidth}
            \centering
            \includegraphics[width=0.8\linewidth]{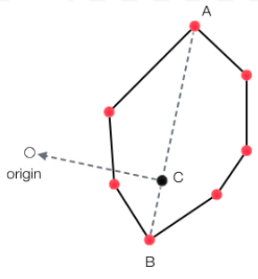}
            \caption{\centering Witness point B is added to $\simplex$. Point C is the closest point to the origin. The direction is $\sd=\overrightarrow{CO}$.}
        \end{subfigure}
        \begin{subfigure}{.45\linewidth}
            \centering
            \includegraphics[width=0.8\linewidth]{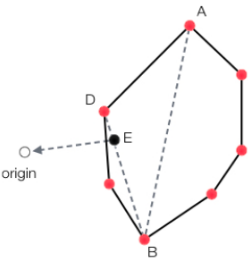}
            \caption{\centering Witness point D is added to $\simplex$. Point E is the closest point to the origin. The direction is $\sd=\overrightarrow{EO}$.}
        \end{subfigure}
        \begin{subfigure}{.45\linewidth}
            \centering
            \includegraphics[width=0.8\linewidth]{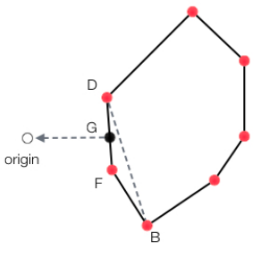}
            \caption{\centering Witness point F is added to $\simplex$. Point G is the closest point to the origin. This is the best simplex.}
        \end{subfigure}
        \caption{\centering A two dimensional example of the GJK algorithm over four iterations.}
        \label{fig:gjk-iter}
        \vskip -0.5cm
    \end{figure}

    \noindent
    At each iteration of the GJK algorithm, a {\em witness point} $\witness \in \{ \bP \ominus \bQ \}$ is chosen, and the set of witness points are used to form a simplex $\simplex$ and the direction from the simplex to the origin $\sd$. The algorithm repeats until (i) $\simplex$ contains the origin, which means the objects are intersecting, or (ii) no better $\simplex$ can be formed. In the latter case, the point on $\simplex$ closest to the origin has as its magnitude the shortest distance between the objects. The method can be visualized in Fig.~\ref{fig:gjk-iter}.
    
    Ideally, the witness point $\witness$ is chosen as the farthest point from the hyperplane defined by the current search direction $\sd$. This is done through a {\em support function}, which finds the point in $\bP\ominus \bQ$ that has the highest scalar product with $\sd$. The function is implemented in a way that  $\bP\ominus \bQ$  does not need to be explicitly computed by applying the {\em support function} on $\bP$ and $\bQ$ by Eq. \ref{eq:proof-support}.
    \begin{equation}
    \begin{aligned}
    \label{eq:proof-support}
    \witness &= \supportFun(\bP\ominus \bQ, \sd)\\
    &= \operatorname*{argmax}_{\bc \in \{\bP\ominus \bQ\}}(\sd \boldsymbol{\cdot} \bc) = \operatorname*{argmax}_{\bp \in \bP, \bq \in  \bQ}(\sd \boldsymbol{\cdot} (\bp - \bq)) \\
        &= \operatorname*{argmax}_{\bp \in \bP, \bq \in  \bQ}((\sd \boldsymbol{\cdot} \bp) - (\sd \boldsymbol{\cdot} \bq))\\
    &= \operatorname*{argmax}_{\bp \in \bP}(\sd \boldsymbol{\cdot} \bp) - \operatorname*{argmin}_{\bq \in \bQ}(\sd \boldsymbol{\cdot} \bq)\\
    &= \operatorname*{argmax}_{\bp \in \bP}(\sd \boldsymbol{\cdot} \bp) - \operatorname*{argmax}_{\bq \in \bQ}(-\sd \boldsymbol{\cdot} \bq)\\
    &= \supportFun(\bP, \sd) - \supportFun(\bQ, - \sd)\\
    \end{aligned}
    \end{equation}

    \noindent
    Given the current simplex $\simplex$, the {\em distance function} finds the point that is closest to the origin and defines the search direction $\sd$. This function computes the barycentric coordinates and a subset from $\simplex$ to identify the edge/face of the simplex that is closest to the origin. Geometrically, $\sd$ is perpendicular to an edge/face of the simplex. After finding the closest point, $\simplex$ is reduced to a set with the witness points corresponding to the point/edge/face on which the closest point lies. In the next iteration, $\witness$ is added to $\simplex$, forming a new simplex. The algorithm stops when $\origin$ is found within $\simplex$ or the generated $\sd$ repeats in which case the \textit{bestSimplex} is found. A generic GJK is summarized in Algorithm~\ref{algo:GJK}.
    
    \SetKwFor{Loop}{Loop:}{}{}
    \begin{algorithm}[ht]
        \SetAlgoLined
        \caption{GJK Distance Algorithm}
        \label{algo:GJK}
        \BlankLine
        \SetKwInOut{Input}{Input}
        \SetKwInOut{Output}{Output}
        \Input{{$\bP$, $\bQ$}}
        \Output{$d$: the shortest distance between $\bP$ and $\bQ$}
        \BlankLine
        $\simplex$ = $\{\emptyset\}$\\
        Initialize $\sd$ (any initial value will give the same result)\\
        \Loop{}{
            $\witness = \supportFun$($\bP\ominus \bQ$, $\sd$)\\
            Add $\witness$ to $\simplex$\\
            $\sd = \distanceFun(\simplex)$\\
            \uIf {$\simplex$ contains $\origin$} {
                \Return $d=0$\\
            }
            \uIf{bestSimplex}{
                \Return $d=\|\sd\|$\\
            }
        }
    \end{algorithm}
    \vskip -0.5cm
    
\subsection{Recursive Gilbert-Johnson-Keerthi Algorithm}
\label{sec:RGJK} 
    \noindent
    The recursive GJK (RGJK) algorithm adds a recursive element to the {\em support function} in order to avoid exhaustive computations~\cite{ong1997GJK}. At each iteration, we start from the point of the previous iteration and check its neighbouring points. The point on the object with the highest scalar product can be found by a hill-climbing technique. This greatly reduces the number of points to verify and the computation time.

\section{Proposed Method}
\label{sec:proposed-method}
\noindent
    The proposed method is split into two parts. First, the techniques used for modelling the robot and workspace are presented, along with the pipeline for generating the models. Afterwards, our variant of the GJK distance sub-algorithm using a geometric approach is described. 

\subsection{Environment Modeling}
\label{sec:model}
\noindent
    In order to be able to use the advantages of the RGJK algorithm, the components in the model of the environment, such as the links of the robot, the objects in the workspace, and the end effector, need to be represented in a way that the neighbours of any given point of the mesh can be known. To gain this feature, the mesh can be imported into a graph data structure, where the vertices of the mesh are the nodes of the graph and the edges of the mesh populate the adjacency list of the nodes of the graph. This representation gives the neighbouring vertices in the mesh through the adjacency list of a node, providing the ability to use the recursive support function of the RGJK algorithm.
    
    Meshes are easily imported from files which have a 3D file format, such as Standard Triangle Language (STL)(*.stl), COLLADA (*.dae), and Wavefront (*.obj). To import a mesh into the collision detection system, a method for converting the 3D geometry contained in a file format to a graph data structure is required. In the case of an STL file, the mesh is described by triangles formed by the vertices, which makes it easy to find neighbouring vertices. The graph representation of the mesh can be built while reading the file, giving a new index for each new vertex, and adding vertices to the adjacency list of vertices that they form triangles with.
    
\begin{figure}[t]
        \begin{subfigure}{.49\linewidth}
            \centering
            \includegraphics[scale=0.27]{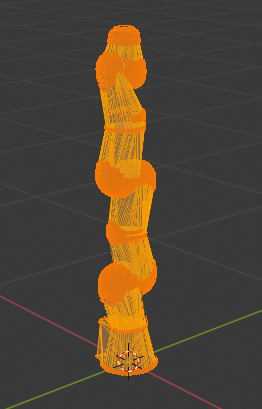}
            \caption{\centering KUKA LBR iiwa 7 R800 Mesh Model}
            \label{fig:mesh-kuka}
        \end{subfigure}
        \begin{subfigure}{.49\linewidth}
            \centering
            \includegraphics[scale=0.154]{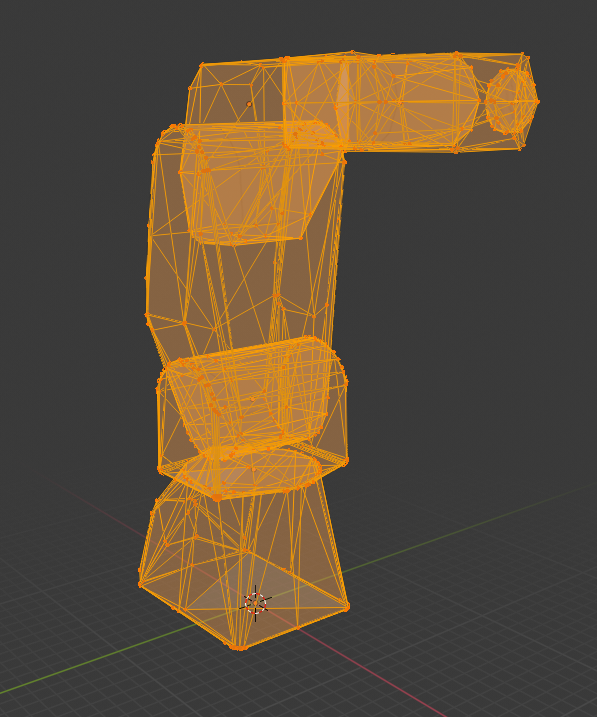}
            \caption{\centering Mecademic Meca500 Mesh Model}
            \label{fig:mesh-meca500}
        \end{subfigure}
        \caption{\centering Mesh Models of the robots used for experimentation taken from their STL files with all joint angles equal to zero.}
        \label{fig:mesh-models}
        \vskip -0.5cm
    \end{figure}

    For each component to be modelled in the environment, the component needs a mesh representation with a frame of reference for the vertices and a transformation matrix that places it in the environment. In the case of the robot links, the reference frame of the mesh representing the link needs to match the reference frame of the forward kinematics of the robot. By matching the reference frames of the links with the reference frames of the forward kinematics of the robot, we model the robot for all its poses as in Fig.~\ref{fig:mesh-models}. 

    Since the robot can move between time steps, the model of the environment needs to be able to update the components of any mobile component in the environment. This is done by having a data structure of graphs representing the components in the environment and only updating the components that can be moved. This saves resources rewriting the same data to stationary components in the environment. For mobile components, a backup of the original graphs generated from the meshes imported from the files are kept in memory, and, for each time step, the new transformations of the mobile components are applied to the meshes and saved into the data structure holding the graphs on the current model. This process is performed once every time step and keeps the model updated with the current information about the environment. The collision detection system using this method of modelling the environment is described in Algorithm~\ref{algo:collision-detection}.

    \begin{algorithm}[h]
        \SetAlgoLined
        \caption{Real-time RGJK Collision Detection}
        \label{algo:collision-detection}
        \BlankLine
        \SetKwInOut{Input}{Input}
        \SetKwInOut{Output}{Output}
        \Input{{$\theta_1$, $\theta_2$, $\dots$, $\theta_N$}: Joint Angles}
        \Output{Detected Collision (Boolean)}
        \BlankLine
        $\bP_1,\bP_2, \dots, \bP_N$ = update the position of all components with {$\theta_1$, $\theta_2$, $\dots$, $\theta_N$} into the same frame\\
        \For{$i=1$ to $N$ } {
            \For{$j=i+1$ to $N$ }{
                    Compute the distance $d_{ij} =$ RGJK($\bP_i$, $\bP_j$) \\ 
            	    \If{$d_{ij} \leq \threshold$}{
            		    \Return True \\
            	    }
            }
        }
        \Return False
    \end{algorithm}
                \vskip -0.5cm

\subsection{The Distance Algorithm}
\label{sec:distAlgo} 
\noindent
    The $distance$ function in Algorithm~\ref{algo:GJK} is where all the heavy lifting of the GJK algorithm is done. This function finds the closest point to the origin on the current simplex $\simplex$, which is also used to find the search direction $\sd$. Previous work used the barycentric coordinates to find this point~\cite{gilbert1988GJK}\cite{cameron1997GJK}\cite{montanari2017GJKSV}, but here, we modified the $distance$ function with a simpler approach from linear algebra.
    
    Since we are using simplexes that only require two to four witness points, we only need to keep these witness points on $\simplex$. 
    It is shown that GJK algorithm monotonically converges to the true convex hull of $\{\bP\ominus \bQ \}$~\cite{montanari2017GJKSV}. Thus, we know that the search direction can only move closer to the solution but not further. We divided the search area into smaller regions, and we can discard regions that are further away from the closest point to the origin at the current iteration.

\begin{figure}[h]
        \centering
        \begin{subfigure}{.45\linewidth}
            \centering
            \includegraphics[width=0.9\linewidth]{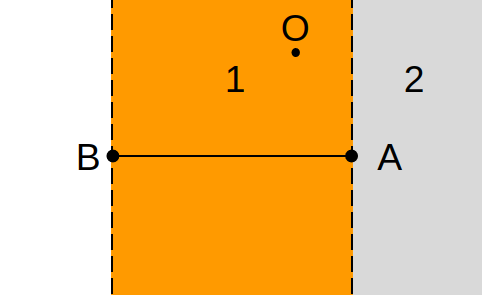}
            \caption{\centering Line simplex regions}
            \label{fig:line-case}
        \end{subfigure}
        \begin{subfigure}{.45\linewidth}
            \centering
            \includegraphics[width=0.8\linewidth]{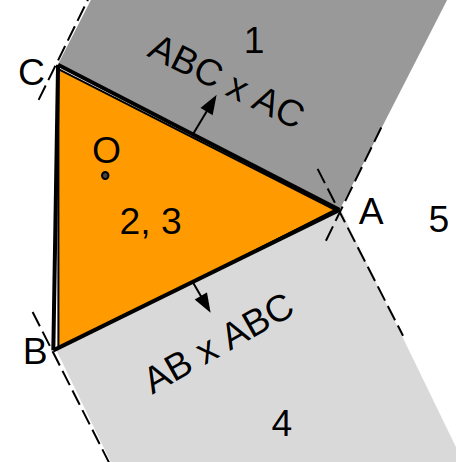}
            \caption{\centering Triangle simplex regions}
            \label{fig:triangle-case}
        \end{subfigure}
        \caption{\centering Regions where the origin can be found.}
        \label{fig:simplex-regions}
                        \vskip -0.5cm
    \end{figure}

    For the line case, there are three regions where the origin can be as seen in Fig.~\ref{fig:line-case}. However, since we know that $A$ is the closest point to the origin in $\simplex$, our approach rules out the 3rd region and only checks regions 1 and 2. If there is a point in region 1 that is the closest point to the origin, then $\overrightarrow{AB} \boldsymbol{\cdot} \overrightarrow{AO} > 0$ is true and the closest point can be found by using the vector rejection formula, $\sd = \overrightarrow{AO} - \frac{\overrightarrow{AB} \boldsymbol{\cdot} \overrightarrow{AO}}{\|\overrightarrow{AB}\|^2} \overrightarrow{AB}$. If $\overrightarrow{AB} \boldsymbol{\cdot} \overrightarrow{AO} > 0$ is false, then the point $A$ is the closest point and we can remove point $B$ from $\simplex$. 

    \noindent
    
    For the triangle simplex, there are 8 regions where the origin could be. Again, using the knowledge that point $A$ is the closest point to date, our approach rules out the regions covered by $B$, $C$ and the line $BC$ as seen in Fig.~\ref{fig:triangle-case}. Our approach only needs to verify 5 regions to find the closest point to the origin. We can verify region 1 first by checking if $(\overrightarrow{ABC} \times \overrightarrow{AC}) \boldsymbol{\cdot} \overrightarrow{AO} >0$ and then if $\overrightarrow{AC} \boldsymbol{\cdot} \overrightarrow{AO} >0$. If these conditions are true, we can treat $A$ and $C$ as a line and go to the line case. We can verify region 4 first by checking if $(\overrightarrow{ABC} \times \overrightarrow{AC}) \boldsymbol{\cdot} \overrightarrow{AO} >0$ and if $ \overrightarrow{AB} \boldsymbol{\cdot} \overrightarrow{AO} >0$, or if $(\overrightarrow{AB} \times \overrightarrow{ABC}) \boldsymbol{\cdot} \overrightarrow{AO} >0$ and if $ \overrightarrow{AB} \boldsymbol{\cdot} \overrightarrow{AO} >0$. This is because we are making a plane without a fixed position. For regions 2 and 3, the verification is simply $ (\overrightarrow{ABC} \times \overrightarrow{AC}) \boldsymbol{\cdot} \overrightarrow{AO} < 0$ and $(\overrightarrow{AB} \times \overrightarrow{ABC}) \boldsymbol{\cdot} \overrightarrow{AO} < 0$. The closest point for this case is computed by using the vector projection of $\overrightarrow{AO}$ on $\overrightarrow{ABC}$, $\sd = \frac{\overrightarrow{ABC} \boldsymbol{\cdot} \overrightarrow{AO}}{\|\overrightarrow{ABC}\|^2}\overrightarrow{ABC}$. We can verify region 5 by ruling out all the other regions.

    For the tetrahedron, there are 15 regions to verify. However, we can reject 7 regions by using the same idea that $A$ is the closest point. We can cover the 8 remaining regions by checking the triangle case of each face with point $A$ as one of its vertices. If the origin is within the tetrahedron, we can make $\sd = \origin$ and return from the RGJK algorithm.
\section{Results}
\label{sec:results}
\noindent
    In this section, the performance of the collision detection system described previously is explored and discussed. The collision detection system, implemented in C, is compared to the Flexible Collision Library (FCL). Both our system described previously and FCL model the robot, apply the transformation to the components of the model and perform the collision checking. Since we wish to be able to use the collision checking in real time, the computation time must be under one millisecond. Both collision systems were used on the KUKA LBR iiwa 7 R800 and the Mecademic Meca500 to compare the flexibility and performance of each system on different models. The simulations only took into account self-collisions, but the system can also be adapted to verify collision in the robot's workspace. Both systems were simulated on a $2.1$~GHz AMD Ryzen 5 pro 3500u processor running Linux. In Algorithm~\ref{algo:collision-detection}, we use $\threshold = 0$.
    
    For the KUKA LBR iiwa, the model was taken from the ROS Industrial Experimental packages for the Kuka manipulators\footnote{\href{https://github.com/ros-industrial/kuka_experimental}{https://github.com/ros-industrial/kuka\_experimental}}. The STL files containing the collision model were processed and converted into convex hull representations. The new convex hulls were used in the experiments.

    For the Mecademic Meca500, the model was taken from Mecademic's ROS package\footnote{\href{https://github.com/Mecademic/ROS}{https://github.com/Mecademic/ROS}}. No pre-processing was required since the meshes in the STL files were already convex hulls.
    
    For both robots, the collision checking was performed on $20000$ different poses. The oriented bounding boxes (OBB) bounded volume hierarchy (BVH) scheme from FCL was used for the comparisons. This scheme was used since OBB is a tight-fitting bounding volume to the underlying geometry and distance querying is said to take longer \cite{pan2012FCL}. Both systems yielded the same conclusions for the collision checking for all the poses, showing that both approaches identify collisions effectively and are functionally equivalent. For the timing performance, our RGJK method is much faster than FCL, where the results are found in TABLE~\ref{table:results-timing}.
    
     \begin{table}[ht]
        \centering
        \begin{tabular}{||c|c|c||}
        \hline\hline
        Robot       & FCL library \cite{pan2012FCL}&   Proposed Method\\
        \hline
        Meca500     & $7.093  \pm 0.649$ & $0.057 \pm 0.012$ \\
        \hline
        Kuka LWR    & $214.935 \pm 76.427 $ &  $0.173 \pm 0.015$\\
        \hline\hline
        \end{tabular}
        \caption{\centering Timing for the self-collision detection averaged over $20 000$ different poses (in milli-seconds)}
        \label{table:results-timing}
    \end{table}
    
    We can see for both robots that the our method is capable of being used in real time while FCL is well over $1$~ms for both cases. Our method took $0.08\%$ of the time FCL needed to compute the self-collision on the KUKA robot and $0.8\%$ of the time FCL needed for the Mecademic robot. One reason for this is the way both approaches model the components in the environment. In our method described previously, the transformation of each component is applied once before starting collision checking. In contrast, in FCL, the transformation needs to be applied on the component each collision check. Our method cuts down time by doing the work once and reusing it for every collision check. 
    
    We also see that our method is better at adapting to more complex models. The method's computation time for performing the self-collision checking on the Mecademic robot increased by $204\%$ when computing the self-collision on the KUKA robot. Using FCL, the increase is of $2930\%$. This is due to the recursive support algorithm making the RGJK method faster by removing computations that do not need to be performed to achieve the same outcome.

\section{Conclusion} 
\label{sec:conclusion}
\noindent
    In conclusion, we proposed a generic method for real-time collision detection.
    The method takes the unified robot description format (URDF) and extends the GJK distance query algorithm for faster computation. Experimental results show that the proposed method is capable of running in real time and outperforms the existing open source library in terms of computation time.
    
    For future work, we will utilize the distance query method developed in this work for real-time collision avoidance systems. Also, we will extend the proposed method to detect collision between the known components (provided by the URDF) and dynamic components (e.g., objects). This can be done by incorporating sensory feedback to gain additional information about the environment in the workspace of the robots.
    
\bibliographystyle{unsrtnat}
\bibliography{references}

\begin{thebibliography}{11}
\providecommand{\natexlab}[1]{#1}
\providecommand{\url}[1]{\texttt{#1}}
\expandafter\ifx\csname urlstyle\endcsname\relax
  \providecommand{\doi}[1]{doi: #1}\else
  \providecommand{\doi}{doi: \begingroup \urlstyle{rm}\Url}\fi

\bibitem[Lin et~al.(2018)Lin, Smith, Babarahmati, Dehio, and
  Mistry]{lin2018projected}
Hsiu-Chin Lin, Joshua Smith, Keyhan~Kouhkiloui Babarahmati, Niels Dehio, and
  Michael Mistry.
\newblock A projected inverse dynamics approach for multi-arm cartesian
  impedance control.
\newblock In \emph{IEEE International Conference on Robotics and Automation},
  pages 1--5, 2018.

\bibitem[Ericson(2004)]{ericson2004real}
Christer Ericson.
\newblock \emph{Real-time collision detection}.
\newblock CRC Press, 2004.

\bibitem[Wu(1992)]{wu1992linear}
Xiaolin Wu.
\newblock A linear-time simple bounding volume algorithm.
\newblock In \emph{Graphics Gems III}, pages 301--306. Elsevier, 1992.

\bibitem[Bergen(1997)]{bergen1997efficient}
Gino van~den Bergen.
\newblock Efficient collision detection of complex deformable models using
  {AABB} trees.
\newblock \emph{Journal of graphics tools}, 2\penalty0 (4):\penalty0 1--13,
  1997.

\bibitem[Chang et~al.(2010)Chang, Wang, and Kim]{chang2010efficient}
Jung-Woo Chang, Wenping Wang, and Myung-Soo Kim.
\newblock Efficient collision detection using a dual {OBB}-sphere bounding
  volume hierarchy.
\newblock \emph{Computer-Aided Design}, 42\penalty0 (1):\penalty0 50--57, 2010.

\bibitem[Ketchel and Larochelle(2006)]{ketchel2006collision}
John Ketchel and Pierre Larochelle.
\newblock Collision detection of cylindrical rigid bodies for motion planning.
\newblock In \emph{IEEE International Conference on Robotics and Automation},
  pages 1530--1535, 2006.

\bibitem[Gilbert et~al.(1988)Gilbert, Johnson, and Keerthi]{gilbert1988GJK}
Elmer~G. Gilbert, Daniel~W. Johnson, and S.~Sathiya Keerthi.
\newblock A fast procedure for computing the distance between complex objects
  in three-dimensional space.
\newblock \emph{IEEE Journal on Robotics and Automation}, 4\penalty0
  (2):\penalty0 193--203, 1988.

\bibitem[Ong and Gilbert(1997)]{ong1997GJK}
Chong~Jin Ong and Elmer~G. Gilbert.
\newblock The {Gilbert-Johnson-Keerthi} distance algorithm: a fast version for
  incremental motions.
\newblock In \emph{IEEE International Conference on Robotics and Automation},
  volume~2, pages 1183--1189, 1997.

\bibitem[Cameron(1997)]{cameron1997GJK}
Stephen Cameron.
\newblock Enhancing {GJK}: computing minimum and penetration distances between
  convex polyhedra.
\newblock In \emph{IEEE International Conference on Robotics and Automation},
  volume~4, pages 3112--3117, 1997.

\bibitem[Montanari et~al.(2017)Montanari, Petrinic, and
  Barbieri]{montanari2017GJKSV}
Mattia Montanari, Nik Petrinic, and Ettore Barbieri.
\newblock Improving the {GJK} algorithm for faster and more reliable distance
  queries between convex objects.
\newblock \emph{ACM Transactions on Graphics (TOG)}, 36\penalty0 (3):\penalty0
  1--17, 2017.

\bibitem[Pan et~al.(2012)Pan, Chitta, and Manocha]{pan2012FCL}
Jia Pan, Sachin Chitta, and Dinesh Manocha.
\newblock {FCL}: A general purpose library for collision and proximity queries.
\newblock In \emph{IEEE International Conference on Robotics and Automation},
  pages 3859--3866, 2012.

\end{thebibliography}

\end{document}